\documentclass[conference]{IEEEtran}
\IEEEoverridecommandlockouts

\AtBeginDocument{%
  \providecommand\BibTeX{{%
    \normalfont B\kern-0.5em{\scshape i\kern-0.25em b}\kern-0.8em\TeX}}}

\usepackage{booktabs}

\usepackage{flushend}
\usepackage{times}  
\usepackage{helvet} 
\usepackage{courier}  
\usepackage{graphicx} 

\usepackage{times}
\usepackage{latexsym}
\usepackage{amsmath}
\usepackage{url}
\usepackage{subcaption}
\usepackage{graphicx}

\usepackage{caption}
\usepackage{multirow}
\usepackage{amssymb}
\usepackage{mathrsfs}
\usepackage{graphics}
\usepackage{graphicx}
\usepackage{ifsym}
\usepackage{amsmath}
\usepackage{subcaption}
\usepackage{booktabs}
\usepackage{multirow}
\usepackage{array}
\usepackage{bbm}

\usepackage{array}
\usepackage{bbm}
\usepackage{multicol}
\usepackage{blindtext}
\usepackage{subcaption}
\usepackage{graphicx}
\usepackage{bm}
\usepackage{multicol}
\usepackage{blindtext}

\usepackage{algorithm}
\usepackage{algpseudocode}
\usepackage{tikz}

\usepackage{lipsum}
\usepackage{makecell}

\begin{document}

\title{A Robust Semantic Frame Parsing Pipeline on a New Complex Twitter Dataset}

\author{Yu Wang and Hongxia Jin}

\maketitle

\begin{abstract}
Most recent semantic frame parsing systems for spoken language understanding (SLU) are designed based on recurrent neural networks. These systems display decent performance on benchmark SLU datasets such as ATIS or SNIPS, which contain short utterances with relatively simple patterns. However, the current semantic frame parsing models lack a mechanism to handle out-of-distribution (\emph{OOD}) patterns and out-of-vocabulary (\emph{OOV}) tokens. In this paper, we introduce a robust semantic frame parsing pipeline that can handle both \emph{OOD} patterns and \emph{OOV} tokens in conjunction with a new complex Twitter dataset that contains long tweets with more \emph{OOD} patterns and \emph{OOV} tokens. The new pipeline demonstrates much better results in comparison to state-of-the-art baseline SLU models on both the SNIPS dataset and the new Twitter dataset\footnote{Our new Twitter dataset can be downloaded from \url{https://1drv.ms/u/s!AroHb-W6_OAlavK4begsDsMALfE?e=c8f2XX}}. Finally, we also build an E2E application to demo the feasibility of our algorithm and show why it is useful in real application.
\end{abstract}

\section{Introduction}
Semantic frame parsing is an important research topic for understanding natural language that aims to construct a semantic frame that captures the semantics of user utterances/queries. In general, simple semantic frame parsing consists of two main tasks \cite{marzinotto2018semantic}: detecting a related event for a given sentence, named a {\it{frame}}, and tagging all the words in the sentence associated with this event, named {\it{frame elements}}. Hence, this technique can also be understood as the combination of a sentence-level classification task and a sequence labeling task. Semantic frame parsing has been widely used for spoken language understanding (SLU) to develop personal AI assistants and chatbots. In SLU, two semantic frame parsing tasks are specified as {\it{intent}} detection and {\it{slot}} filling, which correspond to the {\it{frame}} and {\it{frame elements}}, respectively, 
in a conventional semantic frame parsing task. For a given utterance or sentence, the first task is to label the most likely intent of the utterance, and the second step is to assign a slot label to each of the tokens in the utterance. This information is then combined and further utilized by the downstream components in an SLU system. It is also common practice to jointly train two tasks in a semantic frame parsing system, \emph{i.e.}, to use one model to jointly perform intent classification and slot tagging \cite{zhang2016joint,liu2016attention,xu2013convolutional,goo2018slot,wang2018bi,chen2019bert,lee2018coupled,wang2018new,wang2018deep,wang2019deep,wang2021end,wang2020interactive,wang2020new,wang2020bi,wang2020multi,wang2021adversarial}.

Conventionally, semantic frame parsing can be achieved by a variety of techniques, including conditional random fields (CRFs) \cite{xu2013convolutional,yao2014recurrent}, hidden Markov chains (HMMs) \cite{he2003hidden} and other machine learning models, including decision trees and support vector machines (SVMs) \cite{raymond2007generative}. In recent years, many SLU semantic frame parsing models have been built upon recurrent neural network (RNN) models due to their advantages in performing sequence prediction. Some of these models demonstrate state-of-the-art performance on benchmark SLU datasets, such as ATIS \cite{price1990evaluation} and SNIPS \cite{coucke2018snips}. 
Despite the decent performance of these models, however, it is worth noting that most of the queries in these SLU datasets are relatively short (with an average of 11.4 tokens/query for ATIS and 9.02 tokens/query for SNIPS), and the semantic patterns within each dataset are also very similar. The previous RNN models built on these datasets may not perform well for long sentences containing complex and irregular semantic structures, such as Twitter messages. Due to the large variety of different Twitter writing styles, existing RNN models perform poorly at parsing complex Twitter texts. Technically, there are two main reasons for this poor performance:\\
1. RNN models are known for their poor performance in detecting out-of-distribution (\emph{OOD}) patterns, \emph{i.e.}, irregular patterns that were never included in the training dataset. This issue is very common in Twitter messages containing long sentences, where most of the tokens are either contextual words without any slot labels or auxiliary message content that varies from tweet to tweet.\\
2. Existing RNN models cannot handle out-of-vocabulary (\emph{OOV}) tokens very well, \emph{i.e.}, new tokens introduced during the test that were never included in the training dataset. This is mainly because the word vector dictionary for training cannot cover all tokens that appear in the test set.
\begin{figure*}
\centering
\includegraphics[width=0.88\textwidth,height=0.27\textwidth]{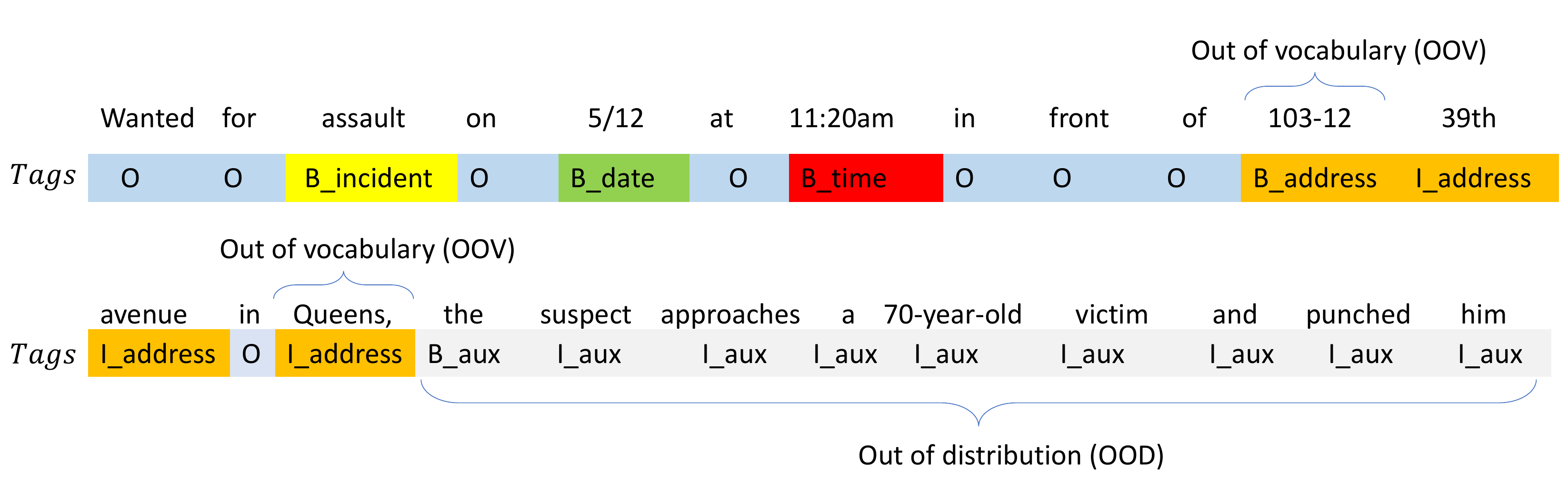}
\caption{ A tweet example in the new twitter dataset}
\label{tweet}
\end{figure*}
Figure \ref{tweet} gives an example of a tweet in the new Twitter dataset to be introduced in this paper; this tweet contains both \emph{OOD} patterns and \emph{OOV} tokens. The newly collected Twitter dataset focuses on incident-related tweets; the main purposes of parsing tasks are to extract important location-related information and detect the incidents mentioned in the selected tweets, such as natural disasters (earthquakes, fires, floods, etc.) and incidents of crime (murder, assault, etc.), so that people can be alerted based on the extracted address and incident information from tweets. From the example given in Figure \ref{tweet}, it can be observed that the tweet contains a description of an auxiliary incident (starting from the label ``B$\_$aux''), which is an \emph{OOD} pattern. It is very unlikely that another tweet can give the same description pattern. Furthermore, the incident date (``5/12'') and the city name in the address (``Queens'') are \emph{OOV} tokens if they are not covered by our word dictionary.

The remainder of this paper is organized as follows:
Section 2 explains all the details concerning the new robust semantic frame parsing pipeline that can handle both \emph{OOD} patterns and \emph{OOV} tokens by leveraging delexicalization on the training dataset and a multistep inference algorithm. In section 3, we introduce a new complex semantic frame parsing dataset containing Twitter messages with incident information; the data collection step, the design of labels, and some dataset statistics are also discussed in this section. We finally perform several SLU semantic frame parsing experiments on both the SNIPS benchmark dataset and the newly introduced Twitter dataset, which are described in section 4. In section 5, we also build an E2E application to demo the feasibility of our algorithm and one possible scenario that can be used as a real application.

\section{A Robust Semantic Frame Parsing Pipeline for handing \emph{OOD} patterns and \emph{OOV} tokens}

In this section, we explain the main algorithm in our semantic frame parsing pipeline employed to handle \emph{OOD} patterns and \emph{OOV} tokens. To achieve these goals, we need to design a semantic frame parsing pipeline by expanding the training dataset using delexicalization and adopting a multistep inference algorithm.

\subsection{Expanding the training dataset}
The first step of the pipeline is to expand our training dataset by:
1. replacing \emph{OOD} patterns using delexicalized tokens and \\
2. substituting \emph{OOV} tokens with the special \emph{$\langle$unk$\rangle$} token.\\
\begin{figure*}[t]
\centering
\includegraphics[width=0.88\textwidth,height=0.52\textwidth]{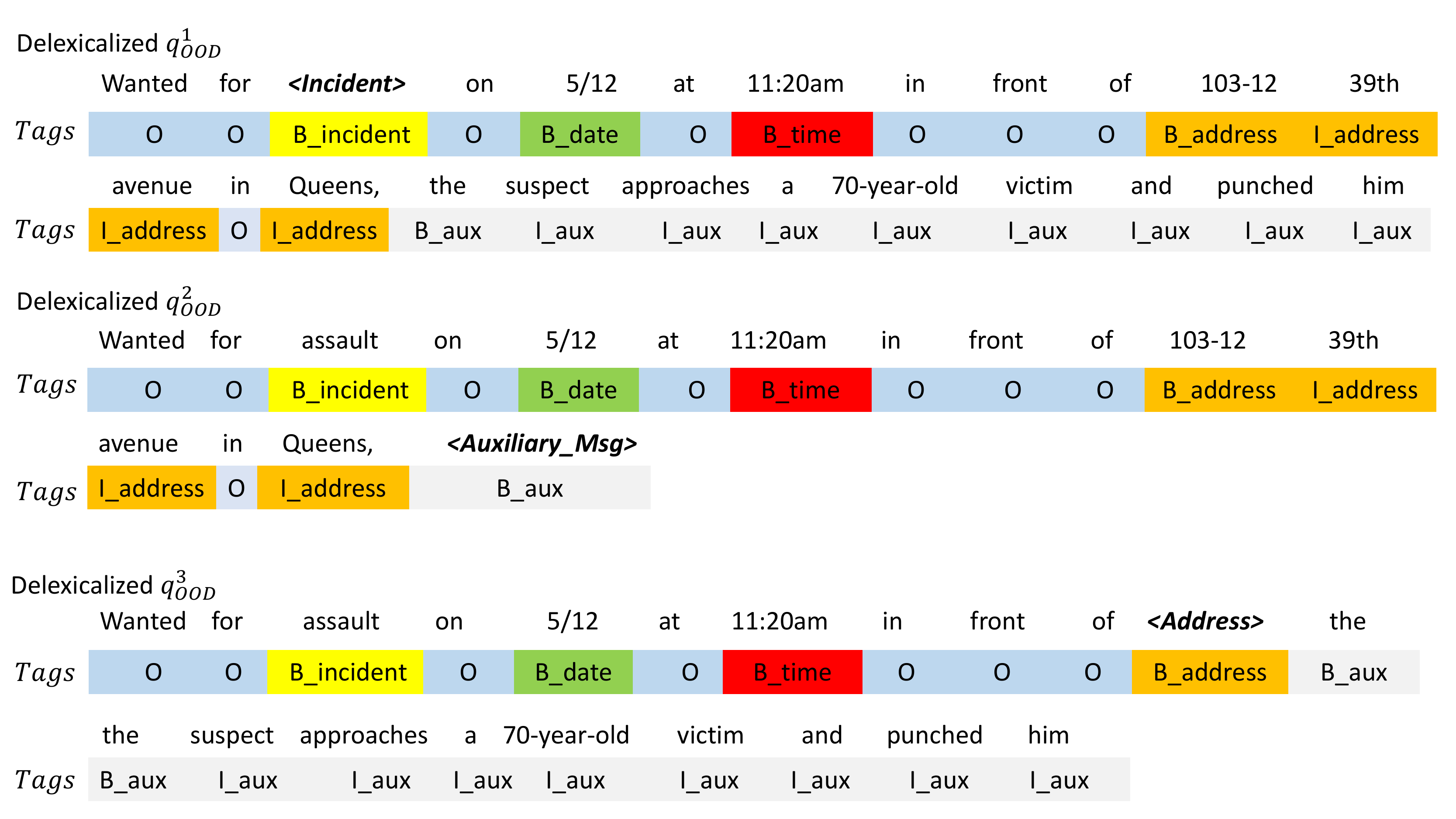}
\caption{Expanding the training set: generating \emph{OOD} varieties}
\label{train1}
\end{figure*}
\begin{figure*}[t]
\centering
\includegraphics[width=0.88\textwidth,height=0.52\textwidth]{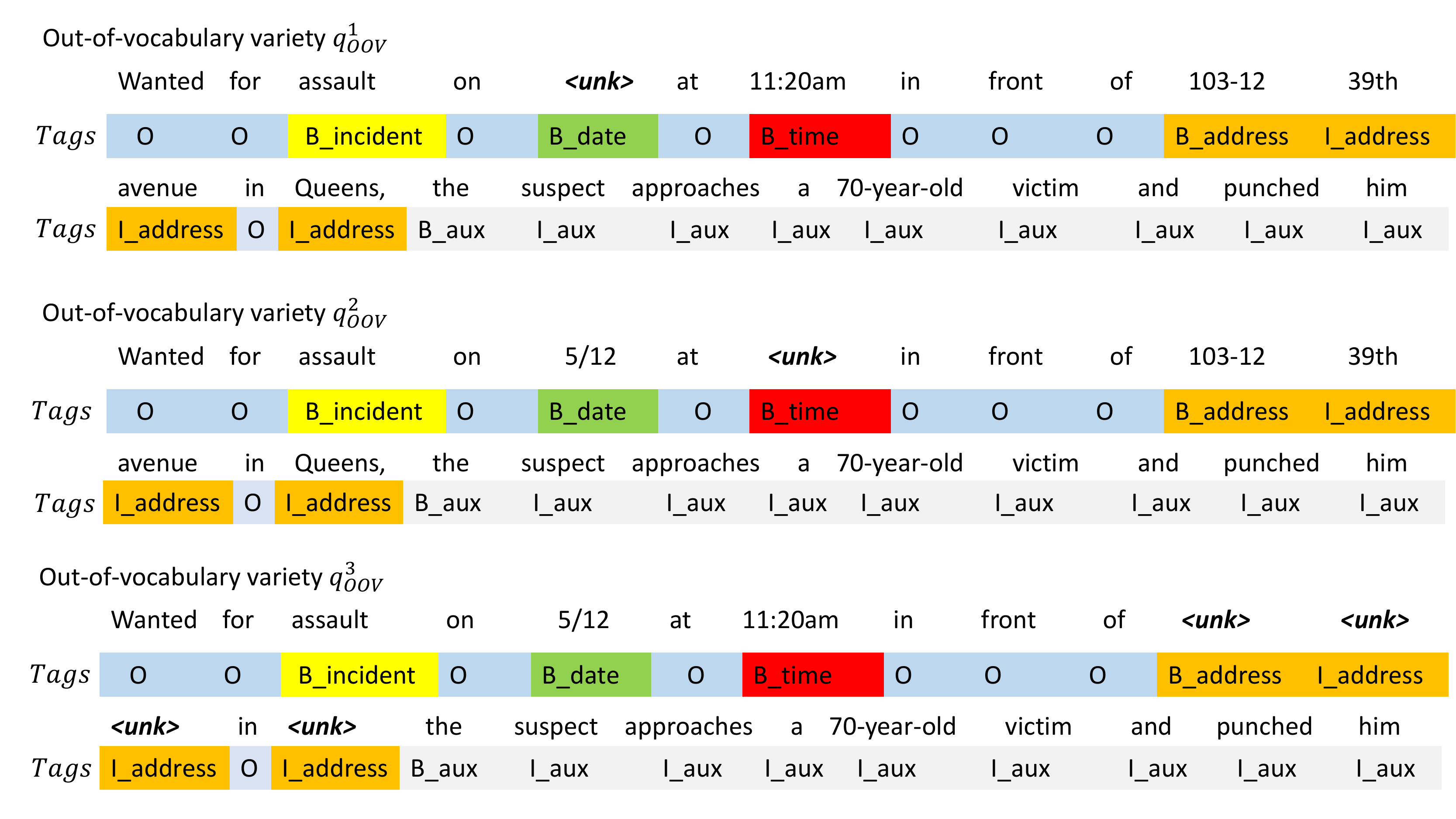}
\caption{Expanding the training set: generating \emph{OOV} varieties}
\label{train2}
\end{figure*}
Figures \ref{train1} and \ref{train2} give examples on how to expand a tweet message $q$ into a set $Q=\lbrace q, q_{ood}, q_{oov} \rbrace$ containing the original sentence $q$, its delexicalized \emph{OOD} varieties $q_{ood}$, and the special token \emph{OOV} varieties $q_{oov}$, which are replaced.
The $q_{ood}$ varieties are generated by delexicalizing the tokens in one of the following ways:\\
1. matching one of our predefined lexicons (such as \emph{$\langle$Incident$\rangle$} in $q_{ood}^1$),\\
2. containing some descriptive text (such as \emph{$\langle$Auxiliary$\_$msg$\rangle$} in $q_{ood}^2$), or \\
3. containing a structured pattern (such as \emph{$\langle$Address$\rangle$} in $q_{ood}^3$).

Each delexicalization step generates one new candidate for $q_{ood}$ to obtain better coverage of different \emph{OOD} varieties.

The $q_{oov}$ varieties are generated by replacing the tokens that contain many other possible varieties but cannot be covered by our training vocabulary; examples of these tokens include times of day ($q_{oov}^1$), dates ($q_{oov}^2$), house numbers, street names, cities in an address ($q_{oov}^3$), etc. To avoid overfitting, we replace these tokens by \emph{$\langle$unk$\rangle$} with a probability of $p_r \in (0,1)$.\\
\emph{Remarks:} \\
1. It is worth noting that the out-of-distribution patterns and out-of-vocabulary tokens in queries represent different scenarios, and they need to be handled separately, which is why two different sets of queries, $q_{ood}$ and $q_{oov}$, are needed to cover both cases.\\
2. It is also observed that some tokens overlap when generating $q_{ood}$ and $q_{oov}$, for example, the delexicalized ``address'' in $q_{ood}^3$ and the unknown replaced token ``address'' in $q_{oov}^3$. We can include both of these tokens in the training data, and there is a merging step if these tokens overlap during the inference step, which is discussed in the next section.

\subsection{A multistep inference algorithm}
Although it is not difficult to delexicalize a sentence in the training dataset based on its given labels, it is not a trival task to delexicalize a sentence during inference since no label is given; \emph{i.e.}, we need to find an approach to delexicalize the test dataset without knowing the groud truth labels. The most common approach for delexicalization during inference (\emph{i.e.}, when the tags are unknown) is based on greedy longest string matching, in which part of the sentence is replaced with special tokens found in the training dictionary \cite{heck2012exploiting,heck2013leveraging,lorenz2002really,mcdonald2011multi}; however, the performance of this approach is not always good, as token replacement does not consider the context of the sentence during inference. Furthermore, the conventional delexicalization approach is not robust and is incompatible with the unknown token \emph{$\langle$unk$\rangle$} defined in our system. Hence, we apply a multistep inference algorithm to generate the best tagging result by taking the context of the sentence and the unknown \emph{OOV} tokens into consideration during delexicalization.

\begin{figure*}[t]
\centering
\includegraphics[width=0.94\textwidth,height=0.28\textwidth]{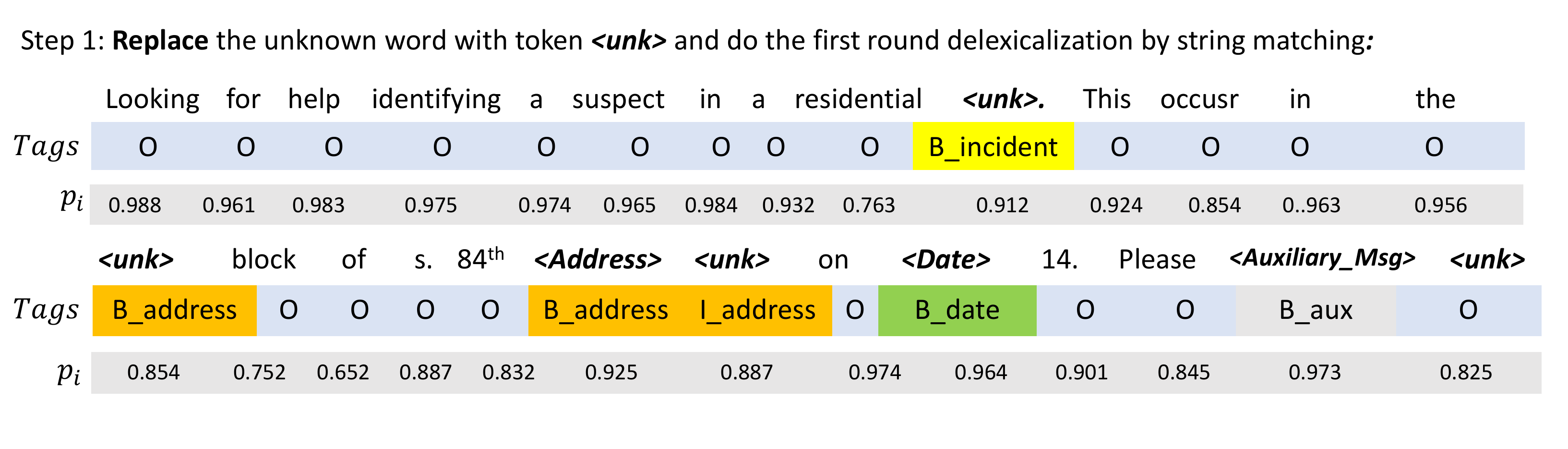}
\caption{Multistep inference algorithm--Step 1: Replace}
\label{replace}
\end{figure*}

\begin{figure*}[t]
\centering
\includegraphics[width=0.94\textwidth,height=0.28\textwidth]{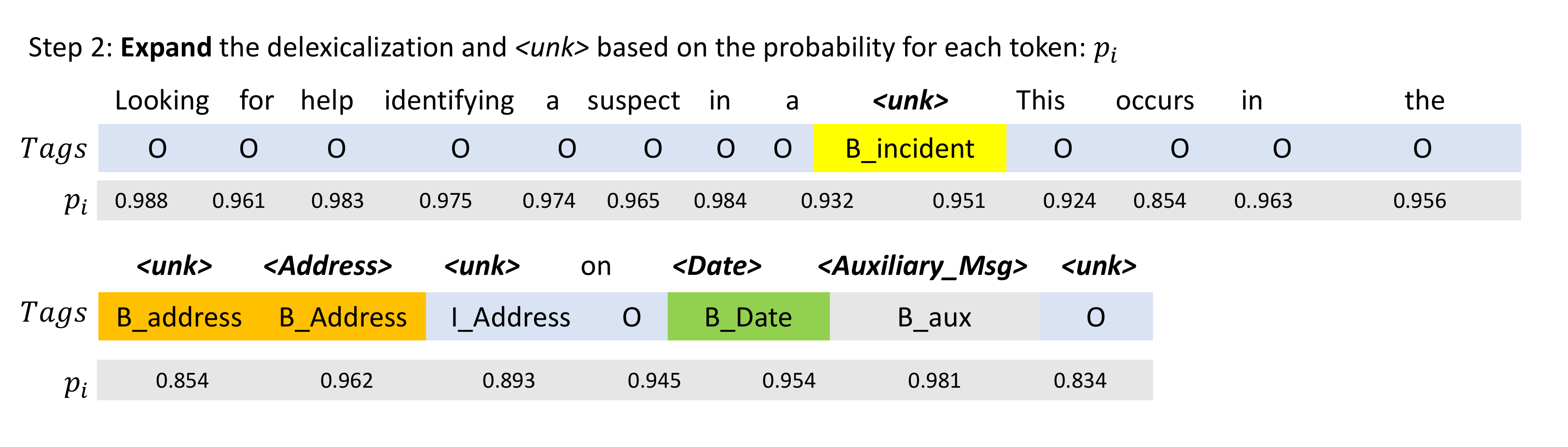}
\caption{Multistep inference algorithm--Step 2: Expand }
\label{expand}
\end{figure*}

\begin{figure*}[t]
\centering
\includegraphics[width=0.94\textwidth,height=0.28\textwidth]{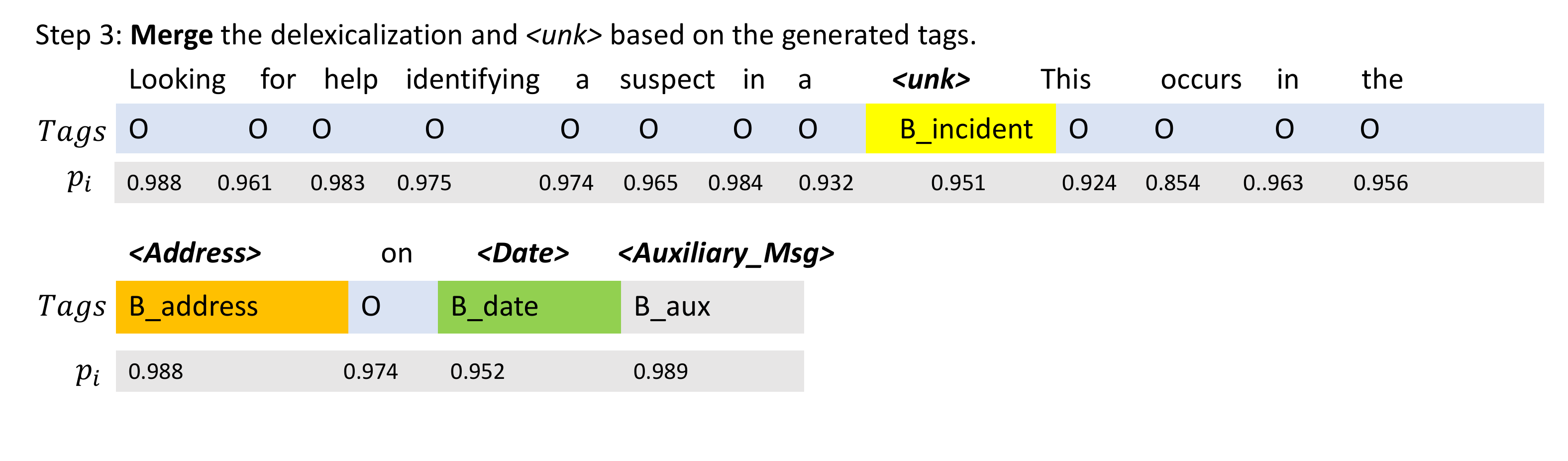}
\caption{Multistep inference algorithm--Step 3: Merge }
\label{merge}
\end{figure*}

Figures \ref{replace}-\ref{merge} demonstrate our multistep inference algorithm using an example in our Twitter test set: ``Looking for help identifying a suspect in a residential burglary. This occurs in the block of S. 84th st., Hazardville on May 14. Please call (860)763-6400''. There are three steps in the inference process:\\
{\bf{Step 1: Replace:}}
The first step is to replace the unknown tokens with a special token, namely, \emph{$\langle$unk$\rangle$}, and then perform the first round of delexicalization through lexicon string matching.
In this example, the words ``burglary'' and ``Hazardville'' are unknown tokens that do not appear in our training data, so these tokens are replaced by \emph{$\langle$unk$\rangle$}. We also delexicalize some words using the tokens \emph{$\langle$Address$\rangle$}, \emph{$\langle$Date$\rangle$}, and \emph{$\langle$Auxiliary$\_$Msg$\rangle$} by matching them with the lexicons collected in the training data.\\ 
{\bf{Step 2: Expand:}}
The second step is to expand the coverage of the delexicalized tokens, such as \emph{$\langle$Address$\rangle$}, \emph{$\langle$Date$\rangle$}, and \emph{$\langle$Auxiliary$\_$msg$\rangle$}, to include more of their neighbor tokens since lexicons generated by the training data cannot cover all the tokens in the test data. To do so, we need to compare the slot tag probability $p_i$ for each token $x_i$ in the given sentence $s$. By defining the left and right boundaries of the delexicalized span as $x_l^{dex}$ and $x_r^{dex}$, respectively, with corresponding tag probabilities of $p_l^{dex}$ and $p_r^{dex}$, the model then compares these probabilities with the tag probabilities of the neighbor tokens, \emph{i.e.}, $p_{l-1}$ for $x_{l-1}$ and $p_{r+1}$ for $x_{r+1}$, to decide whether these neighbor tokens should be included in the delexicalized span. $x_l^{dex} = x_r^{dex} =x^{dex}$ if the span contains only one token.
We expand the delexicalized tokens based on the following rules: If $p_{r+1} < p_r^{dex}$ and $p_{r+1} <Tr$, then delexicalize $x_{r+1}$ by including it into the special delexicalized span, and mark $x_{r+1}$ as the new $x_r^{dex}$. Here, $Tr$ is a predefined threshold probability; we choose $Tr =0.9$ in this paper. Similarly, at the same time step, we check whether $p_{l-1} < p_{l}^{dex}$ and $p_{l-1} <Tr$; then, we delexicalize $x_{l-1}$ by including it into the delexicalized span, and $x_{l-1}$ becomes the new left boundary $x_l^{dex}$.

Then, we rerun the inference model and iteratively perform step 2 and the inference step to expand the delexicalized span until $p_{l-1} > p_{l}^{dex}$ or $p_{l-1} >Tr$ for the left boundary and $p_{r+1} > p_r^{dex}$ or $p_{r+1} > Tr$ for the right boundary.\\
\emph{Remarks:}
The principle responsible for expanding the delexicalized token is as follows: The delexicalization step starts from a single token by replacing the matched tokens with our predefined lexicons. Because the number of predefined lexicons is limited, these lexicons cannot match all the tokens that are supposed to be delexicalized. Hence, this expansion step is added to obtain better token coverage.\\
{\bf{Step 3: Merge:}}
Once the delexicalized spans are expanded and step 2 is finished, the next step is to merge the \emph{$\langle$unk$\rangle$} tokens with the delexicalized spans if they share the same labels. The reason behind the merging step is that there are some unknown tokens that share the same label with the delexicalized tokens during our training data generation step. Hence, it is reasonable to merge the unknown and delexicalized tokens if they share the same labels during inference. We define the \emph{merging rule} as follows:

{\it{If $x_i$ is a delexicalized token $x^{dex}$ with a label $l^{dex}$ and its neighbor token $x_{i-1}$ and/or $x_{i+1}$ is \emph{$\langle$unk$\rangle$} with the same label $l^{dex}$, then the system merges $x_i$ with $x_{i-1}$ and/or $x_{i+1}$ into a single delexicalized token $x^{dex}$. }}

In our example, as shown in Figure \ref{merge}, we merge the three contiguous tokens \emph{$\langle$unk$\rangle$} \emph{$\langle$Address$\rangle$} \emph{$\langle$unk$\rangle$} into one token \emph{$\langle$Address$\rangle$}, as they share the same label (regardless of $B\_$ or $I\_$), \emph{i.e.}, either B$\_$address or I$\_$address. We also merge the two contiguous tokens \emph{$\langle$Auxiliary$\_$msg$\rangle$} \emph{$\langle$unk$\rangle$} into one delexicalized token \emph{$\langle$Auxiliary$\_$msg$\rangle$}.
\section{A new Twitter dataset for incident detection}
In the last section, we use a Twitter example to explain the design details of our semantic frame parsing pipeline. In this section, we attempt to provide all the details about how we collect and label this new Twitter dataset.

There are two main reasons for collecting such a dataset for semantic parsing:\\
1. Currently, most of the benchmark SLU datasets for semantic frame parsing (\emph{e.g.}, ATIS and SNIPS) contain only short spoken utterances. The average query length is relatively short (the average length is 11.4 tokens for ATIS and 9.02 tokens for SNIPS). Hence, existing semantic frame parsing models cannot handle cases in which the utterance is long or there are many out-of-distribution patterns and out-of-vocabulary tokens.\\
2. The current SLU datasets are mostly collected or generated for personal assistance or searching purposes. As verbal language datasets, their structures are relatively simple, and the patterns of the queries are similar; hence, existing state-of-the-art RNN models perform extremely well on these datasets (F1 is always more than $90\%$). In contrast, a semantic frame parsing dataset that contains sentences with complex patterns is lacking, and the data should be closely related to aspects of daily life.

\subsection{Data collection}
Due to the above two reasons, we collect this Twitter dataset containing incident-related queries. The Twitter data are collected from 82 official public Twitter accounts of police departments and fire departments in 23 main cities in the United States (included in the attachment). The collection period ranges from 2019/1/1 to 2019/2/28.
Considering the average sentence length, compared to the averages of 11.4 tokens/query for ATIS and 9.02 tokens/query for SNIPS, the new Twitter dataset contains 33.2 tokens/tweet on average, which is much longer, hence it can be more likely used to train a SLU model which can handle longer sentences compared to the other two benchmark SLU datasets.
\begin{table}[ht]\scriptsize
\parbox{1\linewidth}{
\centering
	\caption{Ratios of different incident types in the Twitter dataset}
	\label{table:incident_types}
	\begin{tabular}{>{\centering\arraybackslash}p{1.2cm}|>{\centering\arraybackslash}p{0.8cm}|>{\centering\arraybackslash}p{0.8cm}|>{\centering\arraybackslash}p{1.9cm}|>{\centering\arraybackslash}p{1.9cm}}
		\toprule
		
		\multirow{1}{*}{\textbf{Type}}&\multirow{1}{*}{\textbf{Fire}} & \multirow{1}{*}{\makecell{\textbf{Crime}}} &\multirow{1}{*}{\makecell{\textbf{Traffic Accident}}} & \multirow{1}{*}{\makecell{\textbf{Natural Disaster}}}\\
		\midrule
		\multirow{2}{*}{} Ratio &29.7$\%$& 27.6$\%$ &28.4$\%$ &14.3$\%$ \\
		
		\bottomrule		
	\end{tabular}
}
\end{table}
Then, we ask our turkers 
to extract all non-duplicated incident-related queries within the collection period, where the incident types are defined in Table \ref{table:incident_types}. There are four types of incidents in our collected dataset: Fire, Crime, Traffic Accident and Natural Disaster. The proportion of each incident type within the dataset is given in Table \ref{table:incident_types}.  The dataset contains a total number of 885 tweets. 

\emph{Remarks:} The incident types in our Twitter dataset are the output of the sentence-level classifier, similar to the ``intent'' in other SLU datasets.

\subsection{Data labeling}
After collecting the data, we provide our judges with the list of labels in Table \ref{table:tag_types} and ask them to tag the selected sentences accordingly. The tag ``$aux$'' is for auxiliary text to explain the details of an incident mentioned in the tweet, and ``$O$'' is intended for the remaining tokens that cannot be labeled by the other tags. The proportions of the tokens under these labels are also given in Table \ref{table:incident_types}.

\begin{table}[ht]\scriptsize
\parbox{1\linewidth}{
\centering
	\caption{Ratios of different tag types in the Twitter dataset}
	\label{table:tag_types}
	\begin{tabular}{>{\centering\arraybackslash}p{1.1cm}|>{\centering\arraybackslash}p{0.75cm}|>{\centering\arraybackslash}p{0.75cm}|>{\centering\arraybackslash}p{0.75cm}|>{\centering\arraybackslash}p{0.8cm}|>{\centering\arraybackslash}p{0.75cm}|>{\centering\arraybackslash}p{0.75cm}}
		\toprule
		
		\multirow{1}{*}{\textbf{Tag Type}}&\multirow{1}{*}{\textbf{Incident}}&\multirow{1}{*}{\textbf{Date}} &\multirow{1}{*}{\textbf{Time}} & \multirow{1}{*}{\makecell{\textbf{Address}}} &\multirow{1}{*}{\makecell{\textbf{Aux}}} & \multirow{1}{*}{\makecell{\textbf{O}}}\\
		\midrule
		\multirow{2}{*}{} Ratio &7.8$\%$& 5.2$\%$ & 7.4$\%$& 12.8$\%$&23.6$\%$ &43.2$\%$ \\
		
		\bottomrule		
	\end{tabular}
}
\end{table}

Based on the statistics, we can observe that the data contain a large proportion of auxiliary text (labeled as $Aux$) that can vary widely from tweet to tweet, which are considered as the \emph{OOD} patterns. As our data are collected from 82 different twitter accounts, the proportion of \emph{OOV} tokens is also large since the training and test datasets can be from different accounts and their word usage can also be entirely different. A comparison between the proportion of \emph{OOD} patterns and \emph{OOV} tokens in both new Twitter dataset and benchmark SNIPS dataset are given in Table \ref{table:OODOOV}.

\begin{table}[ht]\scriptsize
\parbox{1\linewidth}{
\centering
	\caption{Proportion of \emph{OOD} patterns and \emph{OOV} tokens in the new Twitter dataset and SNIP dataset }
	\label{table:OODOOV}
	\begin{tabular}{>{\centering\arraybackslash}p{3.2cm}|>{\centering\arraybackslash}p{1.8cm}>{\centering\arraybackslash}p{1.8cm}}
		\toprule
		
		\multirow{1}{*}{\textbf{Dataset}} & \multirow{1}{*}{\makecell{\textbf{$OOD$ Patterns($\%$)}}} &\multirow{1}{*}{\makecell{\textbf{$OOV$ Tokens($\%$)}}} \\
		\midrule
		\midrule
		
		\multirow{2}{*}{}SNIPS dataset& 5.2&2.7 \\
		\multirow{2}{*}{}New Twitter dataset& 23.6&26.5 \\

		\bottomrule		
	\end{tabular}
}
\end{table}

\section{Experiment}
In this section, we conduct the experiments on two datasets, including the SNIPS dataset and our new Twitter dataset. The SNIPS dataset has a total of 13,784 training utterances and 700 test utterances. For the Twitter dataset, we split 70$\%$ of the collected data (620 tweets) into the training dataset and the remaining 30$\%$ (265 tweets) into the test dataset. The \emph{OOD} slots in the SNIPS dataset are $object\_name$ and $movie\_name$, while the \emph{OOD} slot in our Twitter dataset is $Aux$. As shown in Table \ref{table:OODOOV}, the \emph{OOV} percentage of SNIPS is approximately 2.7$\%$, and that of the Twitter dataset is approximately 26.5$\%$. Similarly, 5.2$\%$ of the SNIPS dataset consists of \emph{OOD} patterns, while more than 23.6$\%$ of the new Twitter dataset is composed of \emph{OOD} patterns by simply considering the tokens with the tag ``$Aux$''. We can see that there is a large difference in the ratios of \emph{OOD} patterns and \emph{OOV} tokens between the two datasets and that it is definitely more difficult to obtain a decent F1 accuracy on the new Twitter dataset. We implement our new pipeline on several state-of-the-art SLU models and compare its performance with the performance without using our expansion of the training dataset and our multistep inference algorithm. The baseline models are the Attention BiRNN model \cite{liu2016attention}, the Slot-Gated BiRNN model \cite{goo2018slot} and the Bi-model BiLSTM model \cite{wang2018bi}, all of which obtain a state-of-art performance on the ATIS and SNIPS SLU datasets.
\begin{table}[ht]\scriptsize
\parbox{1\linewidth}{
\centering
	\caption{Performance of the intent classification and slot tagging tasks on the SNIPS dataset}
	\label{table:snips_comparison1}
	\begin{tabular}{>{\centering\arraybackslash}p{3.5cm}|>{\centering\arraybackslash}p{1.9cm}>{\centering\arraybackslash}p{2cm}}
		\toprule
		
		\multirow{1}{*}{\textbf{Model}} & \multirow{1}{*}{\makecell{\textbf{Intent Accuracy ($\%$)}}} &\multirow{1}{*}{\makecell{\textbf{Slot Tagging F1 ($\%$)}}} \\
		\midrule
		\midrule
		
		\multirow{2}{*}{}Attention BiRNN& 98.0 &90.64 \\
		\multirow{2}{*}{}Attention BiRNN + new pipeline& \textbf{98.85} &\textbf{94.35} \\
		\midrule
		\multirow{2}{*}{}Slot-Gated BiRNN&93.14 &85.26\\
		\multirow{2}{*}{}Slot-Gated BiRNN+ new pipeline&\textbf{97.28}  &\textbf{89.31}\\
		\midrule
		\multirow{2}{*}{}Bi-model BiLSTM &98.85 & 93.52\\
		\multirow{2}{*}{}Bi-model BiLSTM+ new pipeline&\textbf{99.28} &\textbf{96.36}\\	
		\bottomrule		
	\end{tabular}
}
\end{table}

\begin{table}[ht]\scriptsize
\parbox{1\linewidth}{
\centering
	\caption{Performance of the incident (intent) classification and slot tagging tasks on the new Twitter dataset}
	\label{table:twitter_comparison1}
	\begin{tabular}{>{\centering\arraybackslash}p{3.5cm}|>{\centering\arraybackslash}p{1.9cm}>{\centering\arraybackslash}p{2cm}}
		\toprule
		
		\multirow{1}{*}{\textbf{Model}} & \multirow{1}{*}{\makecell{\textbf{Intent Accuracy ($\%$)}}} &\multirow{1}{*}{\makecell{\textbf{Slot Tagging F1 ($\%$)}}} \\
		\midrule
		\midrule
		
		\multirow{2}{*}{}Attention BiRNN& 68.30&63.04 \\
		\multirow{2}{*}{}Attention BiRNN + new pipeline& \textbf{79.25} &\textbf{76.48} \\
		\midrule
		\multirow{2}{*}{}Slot-Gated BiRNN&63.77 &58.86\\
		\multirow{2}{*}{}Slot-Gated BiRNN+ new pipeline&\textbf{72.83}  &\textbf{69.42}\\
		\midrule
		\multirow{2}{*}{}Bi-model BiLSTM &73.21 & 67.12\\
		\multirow{2}{*}{}Bi-model BiLSTM+ new pipeline&\textbf{81.13} &\textbf{78.16}\\
		
		\bottomrule		
	\end{tabular}
}
\end{table}

Several observations can be made based on the experimental results in Table \ref{table:snips_comparison1} and \ref{table:twitter_comparison1}:\\
1. The new robust semantic frame parsing pipeline can improve the performance of the three baseline models on both the SNIPS dataset and the new Twitter dataset.\\
2. Due to the larger proportions of $\emph{OOD}$ patterns and $\emph{OOV}$ tokens in the Twitter dataset, all the models perform worse on the Twitter dataset than on the SNIPS dataset.\\
3. The relative improvement achieved by using the new pipeline on the Twitter dataset is larger than that on the SNIPS dataset; \emph{i.e.}, the system benefits more from the new pipeline if the dataset contains more \emph{OOD} patterns and \emph{OOV} tokens.

In order to further validate the observation 2 and 3 by comparing the new model's performance specifically on the \emph{OOD} patterns and \emph{OOV} tokens, we also extract the tweets containing \emph{OOD} patterns and \emph{OOV} tokens from test datasets, labeled as $q_{ood}$ and $q_{oov}$, and test the model's slot tagging performance only on these data. The results are given in Table \ref{table:snips_comparison2} and Table \ref{table:twitter_comparison2}.
\begin{figure*}[ht]
\centering
\includegraphics[width=1.1\textwidth,height=0.9\textwidth]{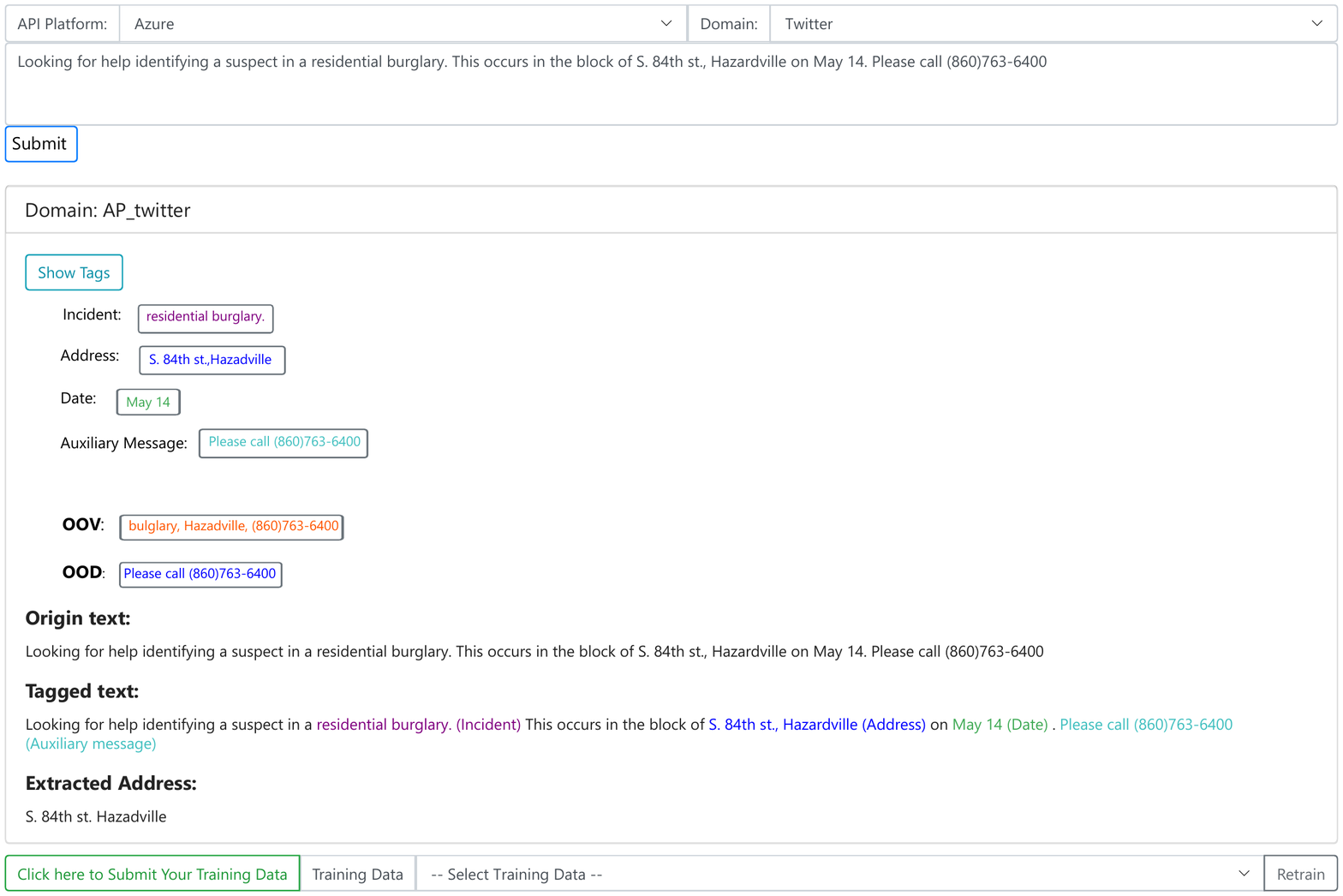}
\caption{Application: A web-based twitter parser}
\label{demo}
\end{figure*}
\begin{table}[ht]\scriptsize
\parbox{1\linewidth}{
\centering
	\caption{Slot Tagging F1 ($\%$) on the data containing \emph{OOD} patterns and \emph{OOV} tokens in SNIPS dataset}
	\label{table:snips_comparison2}
	\begin{tabular}{>{\centering\arraybackslash}p{3.5cm}|>{\centering\arraybackslash}p{1.9cm}>{\centering\arraybackslash}p{2cm}}
		\toprule
		
		\multirow{1}{*}{\textbf{Model}} & \multirow{1}{*}{\makecell{\textbf{F1 of $q_{ood}$ ($\%$)}}} &\multirow{1}{*}{\makecell{\textbf{F1 of $q_{oov}$ ($\%$)}}} \\
		\midrule
		\midrule
		
		\multirow{2}{*}{}Attention BiRNN& 66.12&68.38 \\
		\multirow{2}{*}{}Attention BiRNN + new pipeline& \textbf{87.81} &\textbf{90.35} \\
		\midrule
		\multirow{2}{*}{}Slot-Gated BiRNN&63.53 &65.35\\
		\multirow{2}{*}{}Slot-Gated BiRNN+ new pipeline&\textbf{83.26}  &\textbf{86.83}\\
		\midrule
		\multirow{2}{*}{}Bi-model BiLSTM &68.28 & 70.17\\
		\multirow{2}{*}{}Bi-model BiLSTM+ new pipeline&\textbf{88.35} &\textbf{91.48}\\
		
		\bottomrule		
	\end{tabular}
}
\end{table}

\begin{table}[ht]\scriptsize
\parbox{1\linewidth}{
\centering
	\caption{Slot Tagging F1 ($\%$) on the data containing \emph{OOD} patterns and \emph{OOV} tokens in new Twitter dataset}
	\label{table:twitter_comparison2}
	\begin{tabular}{>{\centering\arraybackslash}p{3.5cm}|>{\centering\arraybackslash}p{1.9cm}>{\centering\arraybackslash}p{2cm}}
		\toprule
		
		\multirow{1}{*}{\textbf{Model}} & \multirow{1}{*}{\makecell{\textbf{F1 on $q_{ood}$ ($\%$)}}} &\multirow{1}{*}{\makecell{\textbf{F1 on $q_{oov}$ ($\%$)}}} \\
		\midrule
		\midrule
		
		\multirow{2}{*}{}Attention BiRNN& 31.13&35.28 \\
		\multirow{2}{*}{}Attention BiRNN + new pipeline& \textbf{62.96} &\textbf{65.73} \\
		\midrule
		\multirow{2}{*}{}Slot-Gated BiRNN&26.52 &30.39\\
		\multirow{2}{*}{}Slot-Gated BiRNN+ new pipeline&\textbf{57.42}  &\textbf{60.04}\\
		\midrule
		\multirow{2}{*}{}Bi-model BiLSTM &36.83 & 40.23\\
		\multirow{2}{*}{}Bi-model BiLSTM+ new pipeline&\textbf{68.95} &\textbf{71.06}\\
		
		\bottomrule		
	\end{tabular}
}
\end{table}

As shown in the tables, the F1 performance improvement by using the new pipeline is more than 20$\%$ on SNIP and over 30$\%$ on the new Twitter dataset. Based on the given performance results, it can be observed that the impact of \emph{OOD} patterns and \emph{OOV} tokens on the new Twitter dataset is much larger than that on SNIPS, hence our new pipeline can give more performance boosts on the Twitter dataset. 

\section{Application}
We utilize the new semantic frame parsing pipeline in our twitter parsing web API, as shown in Figure \ref{demo}. This web portal is designed for parsing tweets. After user types in the tweets, then the system can both extract the tagged slots and identify the \emph{OOV} and \emph{OOD} patterns. The system also allow users to submit their own dataset and retrain their models based on the new submitted dataset. All the user submitted datasets, the trained corresponding parsing models and the training codes are stored on our Azure cloud server \cite{calder2011windows}, the web service framework is written in Flask \cite{grinberg2014flask}, and the web server gateway interface HTTP server is based on Gunicorn \cite{gunicorn2017http}.

As shown by the application, the system can extract and further expose the \emph{OOV} tokens and \emph{OOD} patterns to our users. Based on this information, our users know that what are the missing data pattern and vocabularies needed to be added in their training data, hence can further improve the parser's performance, instead of adding redundant data which are already covered or missing important tokens/patterns those need to be covered.

\section{Conclusion}
In this paper, we design a robust semantic frame parsing pipeline by better incorporating \emph{OOD} patterns and \emph{OOV} tokens. We also introduce a complex Twitter dataset, which contains long sentences with more \emph{OOD} patterns and \emph{OOV} tokens in comparison with earlier benchmark SLU datasets. Our experiments show that the new pipeline can improve the performance of baseline semantic frame parsing models on both the SNIPS dataset and the new Twitter dataset. We also build an E2E application to demo the feasibility of our algorithm and one possible scenario that can be used as a real application.
\newpage
\bibliographystyle{IEEEtran} 
\bibliography{ijcai18MM}
\end{document}